\relax
\documentclass[letterpaper]{article}
\usepackage{named}
\usepackage[colorlinks=true,linkcolor=black,anchorcolor=black,citecolor=black,filecolor=black,menucolor=black,runcolor=black,urlcolor=black]{hyperref}
\usepackage{url}
\usepackage{graphicx}
\usepackage{amssymb}
\usepackage{tabularx,booktabs}
\newcolumntype{Y}{>{\centering\arraybackslash}X}
\newcolumntype{Z}{>{\raggedleft\arraybackslash}X}

\usepackage{xspace}
\usepackage{color}

\newcommand{\placefigure}[2]{%
  \vfill
  \begin{center}
    \includegraphics[scale=#1]{#2}
  \end{center}
  \vfill
}

\usepackage[spanish,english]{babel}

\newcommand{\change}[1]{#1}
\newcommand{\no}[1]{}

\newcommand{\cabbot}{{\sc cabbot}\xspace}
\newcommand{\ribl}{{\sc ribl}\xspace}
\newcommand{\myeq}[1]{%
  \vspace*{0.25cm}
  \centerline{#1}
  \vspace*{0.25cm}
}

\newcommand{\planning}{C}
\newcommand{\observer}{B}

\begin{document}
\title{Domain-independent Generation and Classification of Behavior Traces}
\author{Daniel Borrajo\thanks{On leave from Universidad Carlos III de Madrid. The position at the lab is as a consultant.}
  \hspace*{0.5cm} Manuela Veloso\thanks{On leave from Carnegie Mellon University}\\
  J.P.Morgan AI Research, New York, NY (USA)\\
  \{daniel.borrajo,manuela.veloso\}@jpmchase.com}

\date{}

\maketitle
\begin{abstract}
  Financial institutions mostly deal with people. Therefore, characterizing different kinds of human behavior can
  greatly help institutions for improving their relation with customers and with regulatory offices. In many of such
  interactions, humans have some internal goals, and execute some actions within the financial system that lead them to
  achieve their goals. In this paper, we tackle these tasks as a behavior-traces classification task. An observer agent
  tries to learn characterizing other agents by observing their behavior when taking actions in a given environment. The
  other agents can be of several types and the goal of the observer is to identify the type of the other agent given a
  trace of observations. We present \cabbot, a learning technique that allows the agent to perform on-line
  classification of the type of planning agent whose behavior is observing.  In this work, the observer agent has
  partial and noisy observability of the environment (state and actions of the other agents). In order to evaluate the
  performance of the learning technique, we have generated a domain-independent goal-based simulator of agents. We
  present experiments in several (both financial and non-financial) domains with promising results.
\end{abstract}

\section{Introduction}

Given some training traces obtained by observing at least two kinds of agents, the goal of this research consists of
learning a classifier that can differentiate among those types of agents by observing traces of their behavior. We
assume there is a, usually hidden, rationale for the behavior of agents when taking actions in the environment that
depends on some (again hidden) goals and the states they encounter while taking actions to achieve those goals. And we
also assume goals, states and actions can be represented using standard planning representation languages.

We leverage on previous work on sequence classification in contexts where there was no domain model and the
representation of traces was a vector of features~\cite{xing2010brief}. Some of those approaches did a manual definition
of the relevant features to be used in the classification, which usually resulted in domain-dependent approaches. And
none of these approaches used a relational representation of data in the form of goals, states and actions. Instead, we
assume the other agents use a hidden planning model and the relevant aspects to make the classification depend on the
actions executed and the related states.

Given the setup of an observer agent and a planning-execution agent, several decision-making tasks can be
defined. \change{Within this setting,} most works in automated planning have focused on goal/plan recognition, where the
observer has to infer the goals the planning agent is pursuing~\cite{ramirez10} or the plan it is using to achieve some
goals~\cite{avrahami-zilberbrand05fast}. Once the goals/plans are recognized, other planning-related tasks can be solved
such as generating plans to stop an opponent to reach its goals~\cite{ijcai18-counterplanning} or change the environment
to improve the goal recognition task~\cite{Keren14}. Other uses of traces include learning action
models~\cite{AinetoCO19} or predicting the next action or sequence of actions another agent is going to
perform~\cite{bernard2019accurate,Tax_2017}. However, as far as we know, the sequence classification task has not been
addressed yet within the planning community.

Even if it has been less studied than related tasks in the context of automated planning, many real-world tasks benefit
directly from this research. Some of these domains have been studied in the context of domain-dependent
approaches~\cite{xing2010brief}. Examples are: predicting whether someone will buy a product from the web clicks
sequence; detecting intrusions in network or stand-alone computer systems; classification of anomalous behavior in
public spaces (e.g. terrorism); machines monitoring the behavior of other machines; or labeling an opponent's behavior
in a game. In the case of financial applications there are numerous examples of the use of this task such as: fraud or
anti-money laundering detection; classifying malicious traders; attrition prediction; offering new services to
customers; or detection of users that will complain.

We present as contributions: a learning technique that can classify in agents' types based on their behavior expressed
in observation traces; and a domain-independent simulator of agents' behavior based on dynamic goal generation, planning
and execution. We name the first contribution Classification of Agents' Behavior Based on Observation Traces
(\cabbot). Some of the simulator features are: explicit reasoning on goals generation, modification and removal; ability
to inject new instances when needed; several methods for generating goals (goals schedule, behavior-based random
generation); exogenous events; non deterministic execution of actions; and partial and noisy observability.

A version of this work was published in~\cite{icaif20}. We focused before on its application to money laundering, while
this paper focuses on its general applicability to planning tasks. Therefore, the description of the techniques and the
simulator are centered on the underlying planning tasks, and the experiments report on several domains. Thus, we also
present as contribution several domains designed for this task, whose detailed description is included in the
experimental section. The domains range from a simplified terrorist domain to a service cars domain and two financial
services related ones.  The results show that \cabbot can accurately classify agents in those domains.

\section{Background}

Given that we assume agents' rational behavior to be based on the concepts of goals, states and actions, we will use the
automated planning formalism to describe the tasks we deal with in
this paper~\cite{planning-book}.

\subsection{Automated Planning}

We use the standard classical {\sc strips} definition of a planning task, augmented with numeric variables
(functions). A planning task is defined as $\Pi=\langle F,A,I,G\rangle$, where $F$ is a set of boolean and numeric
variables, $A$ is a set of actions, $I\subseteq F$ is the initial state and $G\subseteq F$ is a set of goals. Each
action $a\in A$ is defined in terms of its preconditions (pre($a$)) and effects (eff($a$)).  Effects can set to true the
value of a boolean variable (add effects, add($a$)), set to false the value of a boolean variable (del effects,
del($a$)), and change the value of a numeric variable (numeric effects, num($a$)). We will denote with $S$ the set of
all states. A (full) state is a valuation of all the variables in $F$; a boolean value for all the boolean variables and
a numeric value for the numeric ones. Action execution is defined as a function $\gamma: S,A\rightarrow S$; that is, it
defines the state that results of applying an action in a given state. It is usually defined as
$\gamma(s,a)=(s\setminus$del($a$))$\cup$add($a$) if pre($a$)$\subseteq s$ when only boolean variables are
considered. When using numeric variables, $\gamma$ should also change the values of the numeric variables (if any) in
num($a$), according to what the action specifies; increasing or decreasing the value of a numeric variable or assigning
a new value to a numeric variable. If the preconditions do not hold in $s$, the state does not change.

The solution of a planning task is called a plan, and it is a sequence of instantiated actions that allows the system to
transit from the initial state to a state where goals are true. Therefore, a plan
$\pi=\langle a_1,a_2,\ldots a_n\rangle$ solves a planning task $\Pi$ (valid plan) iff $\forall a_i\in\pi, a_i\in A$, and
$G\subseteq\gamma(\ldots\gamma(\gamma(I,a_1),a_2)\ldots),a_n)$. In case the cost is relevant, each action can have an
associated cost, $c(a_i), \forall a_i\in A$ and the cost of the plan is defined as the sum of the costs of its actions:
$c(\pi)=\sum_i c(a_i), \forall a_i\in\pi$.

The planning community has developed a standard language, PDDL (Planning Domain Description Language), that allows for a
compact representation of planning tasks~\cite{PDDL}. Instead of explicitly generating all states of $\Pi$, a lifted
representation in a variation of predicate logic is used to define the domain (predicates and actions) and the problem
to be solved (initial state and goals).
%% If we consider a financial domain where users perform monetary transactions,
%% Figure~\ref{fig:domain} shows an example of the action {\tt money-transfer} modeled in PDDL.\footnote{For the sake of
%%   clarity, we have omitted other elements of the actual action model.} Each action is described in terms of its name, a
%% list of parameters (and the type of each parameter), a logic formula describing the preconditions of the action to be
%% executed, and a formula describing what effects are expected after executing the action in the environment. In this
%% case, the preconditions establish that in order to make a money-transfer with a given amount, there should be enough
%% balance in the origin account, and the destination account should be different that the origin account. Once the
%% transaction is performed, the balance of both accounts is updated and the information on the transaction is also
%% updated.

%% \begin{figure}[hbt]
%% \begin{verbatim}
%% (:action money-transfer
%%  :parameters (?ac - account ?t - transfer
%%               ?ad - account)
%%  :precondition (and (>= (balance ?ac)
%%                         (transaction-amount ?t))
%%                     (not (= ?ac ?ad)))
%%  :effect (and (transaction-origin ?t ?ac)
%%               (transaction-destination ?t ?ad)
%%               (decrease (balance ?ac)
%%                         (transaction-amount ?t))
%%               (increase (balance ?ad)
%%                         (transaction-amount ?t))))
%% \end{verbatim}
%% \caption{Model of the action {\tt money-transfer} in the AML domain.}
%% \label{fig:domain}
%% \end{figure}

\subsection{Multi-Agent Framework}

In this work we consider at least two agents: acting agent, $\planning$ (e.g. bank customer) and observer agent,
$\observer$ (e.g. financial institution or bank). In order to create a realistic environment, we will consider that they
have different observability of the environment. Thus, each one of them will have its own definition of a planning task,
as it has already been defined in cooperative~\cite{TorrenoOKS17} and adversarial~\cite{ijcai18-counterplanning}
multi-agent settings. In the case of $\planning$, its planning task can be defined as
$\Pi_\planning=\langle F_\planning,A_\planning,I_\planning,G_\planning\rangle$. In the case of $\observer$, we do not
consider here its ability to plan.

$\observer$ has a partial (public) view of $\planning$'s task. This view can be defined as
$\Pi_{\observer,\planning}=\langle
F_{\observer,\planning},A_{\observer,\planning},I_{\observer,\planning},\emptyset\rangle$, where
$F_{\observer,\planning}\subseteq F_\planning$, $A_{\observer,\planning}\subseteq A_\planning$,
$I_{\observer,\planning}\subseteq I_\planning$ \change{and the goals are unkown, represented as $\emptyset$}. This view
corresponds to the public part of those variables in other Multi-Agent Planning works~\cite{TorrenoOKS17}. It also has a
partial view of the initial state and the actions; since there will be some actions executed by $\planning$, or some
preconditions or effects of those actions that $\observer$ will not observe. $\observer$ has no observability of
$\planning$'s goals. This assumption contrasts with goal and planning recognition work that assumes a set of potential
goals are known~\cite{ramirez10}. In our case, this set would amount to all possible goals that can be defined given a
domain (infinite in most cases). \change{Finally, we relax previous works' requirement on $\planning$ rationality;
  $\planning$ can generate optimal or sub-optimal plans.}

As an example, a customer might have goals that are not observed by the financial institution, such as having committed
a crime, or laundered money. Other goals will be observable only after the customer has executed actions within the
financial system that might reveal them, such as having opened an account, worked for a company, made a money transfer,
or withdrawn money from a bank. In relation to states, there will be information known by the customer that is not
observable by the financial institution, such as how many hours the customer works, or products bought using
cash. Similarly, some information will be known, such as products or services bought using financial instruments of the
corresponding financial institution, or bills payed to utility companies. Finally, there will be actions performed by
the customer that will not be observed by the financial institution, such as committing a crime, while others will be
observable, such as making a money transfer.

Once $\planning$ starts generating plans and executing the actions on those plans, $\observer$ will be able to see: if
the actions in $A_{\observer,\planning}$ are executed; and the components of the state related to variables in
$F_{\observer,\planning}$. A planning trace $t_\planning$ is a sequence of states and actions executed by $\planning$ in
those states:

\[t_\planning=(I_\planning,a_{1},s_{1},a_{2},s_{2},\ldots,s_{n-1},a_{n},s_{n})\]
  
where $s_{i}\in S_{\planning}, a_{i}\in A_{\planning}$. An observation trace is also a sequence of states and actions of
$\planning$ from the point of view of $\observer$
$t_{\observer,\planning}=(I_{\observer,\planning},a'_{1},s'_{1},a'_{2},s'_{2},\ldots,s'_{n-1},a'_{n},s'_{n})$,
where $s'_{i}\in S_{\observer,\planning}, a'_{i}\in A_{\observer,\planning}$.  Each state $s'_{i}$ corresponds to the
partial observability of $\planning$'s state $s_{i}$ by $\observer$. Also, each action $a'_{i}$ corresponds to either an
action that can be observed from $\planning$, $a_{i}$, or a ficticious {\tt no-op} action if $a_{i}$ cannot be observed
by $\observer$.  There is no actual need of requiring the states to be part of the observation; given that $\observer$
has a model of $\planning$'s domain, $\observer$ can always reproduce the corresponding observable states, by simulating
the execution of the observable actions. We will call $T_{\observer,\planning}=\{t_{\observer,\planning}\}$ the set of
traces of agent $\planning$ observed by agent $\observer$.

%% Since that there will be at least two types of planning (observed) agents which we want \cabbot to identify,
In the classification task we are addressing in this paper, there are two $\planning$ agents we would like the learning
system to differentiate by observing their behavior traces. As an example, consider a criminal and a regular customer.
We want to address non-trivial learning tasks.  Therefore, we assume there is nothing in the observable state that
directly identifies one or the other type of $\planning$ agent. Nor there is any difference on the observable actions
between the ones that can be executed by one or the other type of $\planning$. Formally, given two different types of
$\planning$, $\planning_1$ and $\planning_2$, $\observer$'s observable information on both should be the same:

\[\Pi_{\observer,\planning_1}=\Pi_{\observer,\planning_2}=\langle F_{\observer,\planning_1},A_{\observer,\planning_1},I_{\observer,\planning_1},\emptyset\rangle\]

\section{Learning to Classify Behavior}

%% Once we have presented how $P$ generates traces of observations by integrating planning, execution and meta-reasoning
%% capabilities, we will now focus on $O$.
$\observer$'s main task consists of learning to classify among the different types of $\planning$ (behaviors). The
learning task can be defined as follows:

\begin{itemize}
\item Given: (1) a set of classes of behavior (labels)
  ${\cal \planning}=\{\planning_1,\planning_2,\ldots,\planning_n\}$; (2) a set of labeled observed traces
  $T_{\observer,\planning_i}, \forall \planning_i\in {\cal \planning}$; and (3) a partially observable domain model of
  each $\planning_i$ given by $\Pi_{\observer,\planning_i}$
\item Obtain: a classifier that takes as input a new (partial) trace $t$ (with unknown class) and outputs the predicted
  class
\end{itemize}

A main requirement of \cabbot is to be domain-independent. Therefore, we will not use any hand-crafting of features for
the learning task. Another characteristic of this learning task is that it works on unbounded size of the learning
examples. Traces can be arbitrarily large, as well as states within the trace and action descriptions (both in the
number of different action schemas, and grounded actions). There is no a priori limit on these sizes. Using fixed-sized
input learning techniques can be difficult in these cases and some assumptions are employed to handle that
characteristic. Hence, we will consider here only relational learning techniques~\cite{dzeroski2010relational}, and, in
particular, relational instance-based approaches~\cite{ribl}.
%% We assume that a small number of training traces will be needed to correctly classify new
%% instances, so the classification cost associated with these techniques is reduced.
Relational learning techniques have been extensively used in the past to learn control knowledge~\cite{jetai}, or
planning policies~\cite{YoonJMLR08,paa12}, among other planning tasks~\cite{kereview-ml}. But, as far as we are aware
of, they have not been used for this learning task.

The key parameter of these techniques is the relational distance between two traces,
$d: T\times T\rightarrow \mathbb{R}$.  In order to define the distance between two traces, $t_1$ and $t_2$, we have
several alternatives.

\begin{itemize}
  \item %% Since traces $t_1$ and $t_2$ can have different lengths, we will only compare a number of elements corresponding to the
%% minimum length of the two traces, $m$. \todo{probably, we do not need to constrain to $m$ given this distance} Let us
%% consider that $t_i^m$ is the trace composed of the last $m$ elements of $t_i$.
    Compute a distance between the sets of actions on each trace. A simple, yet effective, distance function consists of
    using the inverse of the Jaccard similarity function~\cite{Jaccard} as:

\myeq{$d_a(t_1,t_2)=1-\frac{|an(t_1)\cap an(t_2)|}{|an(t_1)\cup an(t_2)|}$}
%% \[d_a(t_1,t_2)=\frac{|a(t_1^m)\cap a(t_2^m)|}{|a(t_1^m)\cup a(t_2^m)|}\]

where $an(t_i)$ is the set of actions' names in $t_i$. This distance is based on the ratio of common action names in
both traces to the total number of different action names in both traces.
%% The original distance metric is applied to sets, so it is insensitive to changes
%% in the sequence of actions. \todo{thus, I am not sure this is the right decision. There are others that are applied to
%%   sequences, as the ones based on edit distances, such as Levenshtein edit distance metric.}

%% We can compute this number comparing the action schemas (lifted actions -- names of the actions--) or by comparing the
%% fully grounded actions. In the latter case, distances will be very close to zero, since the probability of two different
%% traces to contain the same instantiated actions will be very low in most domains. An alternative consists of doing a
%% relational comparison by using unification of two equal action schemas in both traces and then propagating that
%% unification over the remaining traces.  This alternative is the more expensive one, but could yield more accurate
%% results.  \todo{I will have to explain this better and see if it makes sense}

\item Compute distances between sequences of states differences. Given two consecutive states $s_1$ and $s_2$ in a
  trace, we define their associated difference or delta, that represent the new literals in the state after applying the
  action. They are defined as: $\delta_{s_i,s_{i+1}}=s_{i+1}\setminus s_i$.
%%   \no{These subgoals are a superset of the true subgoals, since
%%     actions might have side effects that are not needed later on. We could be more precise and compute only those that
%%     are needed for later actions, but we might end up with null goals, since most goals will not be observable nor the
%%     actions that achieve the true goals.}
  We can compute a distance between the sets of deltas on each trace by using the Jaccard similarity function as before.

  \myeq{$d_\Delta(t_1,t_2)=1-\frac{|\Delta(t_1)\cap \Delta(t_2)|}{|\Delta(t_1)\cup \Delta(t_2)|}$}

  where $\Delta(t_i)=\{\delta_{s_j,s_{j+1}} \mid \forall s_j, s_{j+1}\in t_i, 0\leq j\leq n-1\}$ is the set of deltas of
  a trace $t_i$. Again, we only use the predicate and function names.
  
%% \item use a similar approach. However, apart from having a sequence of states, each state is a set of valuations to
%%   variables.
%% %% And, again the number of variables in two states of the same or
%% %% different traces might not be the same. So, in this case,
%% The overall distance between two traces will be the minimum of the distances of all states in those traces, where each
%% individual distance between two states will be computed as in the case of actions.
%% 
%% \[d_s(t_1,t_2)=\min_i d_f(s_i(t_1),s_i(t_2))\]
%% 
%% \[d_f(s_1,s_2)=\frac{|f(s_1)\cap f(s_2)|}{|f(s_1)\cup f(s_2)|}\]
%% 
%% %% \[d_s(t_1,t_2)=\sum_{i=n-m}^m d_f(s_i(t_1),s_i(t_2))\]
%% %% 
%% %% \[d_f(s_1,s_2)=\frac{|f(s_1^r)\cap f(s_2^r)|}{|f(s_1^r)\cup f(s_2^r)|}\]
%% 
%% where $f(s)$ means the value of variables in state $s$.
%% %% and $r$ is the minimum length of both states. \todo{Again, I am
%% %% not sure about computing this $r$}

%% \item compute the distance between two traces as the combination of two distances, one for actions in the trace, $d_a$
%%   and another one for the deltas $d_\Delta$:
%% 
%%   \[d(t_1,t_2)=d_a(t_1,t_2)\oplus d_\Delta(t_1,t_2)\]
%% 
%%   We can define different combination functions, such as sum, min, weighted sum, \ldots We use the sum in the
%%   experiments.
\item The two previous distances only consider actions and deltas as sets. If we want to improve the distance metric, we
  can use a frequency-based approach (equivalent to an $n$-grams analysis with $n=1$). Each trace is represented by a
  vector. Each position of the vector contains the number of times an observable action appears in the trace. The
  distance between two traces, $d_g$, is defined as the squared Euclidean distance of the vectors representing the
  traces. As before, a new trace is classified as the class of the training trace with the minimum distance to the new
  trace.

\item Instead of using only counts, the distance function can also consider actions and state changes as relational
  formulae and use more powerful relational distance metrics. We have defined a version of the \ribl relational distance
  function~\cite{ribl} adapted for our representation of traces, $d_r$. We needed to adapt it given the different
  semantics of the elements of the traces with respect to generic \ribl representation of examples. Given two traces, we
  first normalize the traces by substitution of the names of the constants by an index of the first time they appeared
  within a trace.  For instance, given the following action and state pair:

    \begin{tabbing}
      $\langle$ \={\tt create-account(customer-234,acc-345)},\\
      \> {\tt \{}\={\tt acc-owner(customer-234,acc-345),}\\
      \> \>      {\tt balance(acc-345)=2000\}} $\rangle$
    \end{tabbing}

    the normalization process would convert the trace to:

    \begin{tabbing}
      $\langle$ \={\tt create-account(i1,i2)},\\
      \> {\tt \{acc-owner(i1,i2)}, {\tt balance(i2)=2000\}} $\rangle$
    \end{tabbing}

    This process allows the distance metric to partially remove the bias related to using different constant names in
    the traces. The distance $d_r$ is then computed as:

    \[d_r(t_1,t_2)=\frac{1}{2}(d_{ra}(t_1,t_2)+d_{r\Delta}(t_1,t_2))\]
%%   \myeq{$d_r(t_1,t_2)=\frac{1}{2}(d_{ra}(t_1,t_2)+d_{r\Delta}(t_1,t_2))$}

  i.e. as the average of the sum of $d_{ra}$ (distance between the actions of the two traces) and $d_{r\Delta}$
  (distance between the deltas of both traces). $d_{ra}$ is computed as: 

  \myeq{$d_{ra}(t_1,t_2)=\frac{1}{Z} \sum_{a_i\in a(t_1)} \min_{a_j\in a(t_2)} d_{f}(a_i,a_j)$}
  
  where $a(t_i)$ is the set of ground actions in $t_i$, $d_f$ is the distance between two relational formulas and $Z$ is
  a normalization factor ($Z=\max\{|a(t_1)|,|a(t_2)|\}$). We normalize by using the length of the longest set of actions
  to obtain a value that does not depend on the number of actions on each set, so distances are always between 0 and
  1. $d_f$ is 1 if the names of $a_i$ and $a_j$ differ. Otherwise, it is computed as:

  \myeq{$d_{f}(a_i,a_j)= 0.5-0.5\frac{1}{|\mbox{arg}(a_i)|} d_{\mbox{arg}}(a_i,a_j)$}

%% \sum_{\mbox{arg}(a_i),\mbox{arg}(a_j)} d_{\mbox{arg}}(\mbox{arg}(a_i),\mbox{arg}(a_j))  

  where $d_{\mbox{arg}}(a_i,a_j)$ is the sum of the distances between the arguments in the same positions in both
  actions. Each distance will be 0 if they are the same constant and 1 otherwise. Again, we normalize the values for
  distances. Also, when two ground actions have the same action name, we set a distance of at most 0.5. For instance, if
  $l_1=${\tt create-account(i1,i2)} and $l_2=${\tt create-account(i3,i2)},
  
  \myeq{$d_f(l_1,l_2)=0.5-0.5\frac{1}{2}(1+0)=0.25$.}

  As a reminder, each trace contains a sequence of sets of literals that correspond to the delta of two
  states. Therefore, $d_{r\Delta}$ is computed as the distance of two sets of deltas of literals ($\Delta(t_1)$ and
  $\Delta(t_2)$). We use a similar formula to the previous ones:

  \myeq{$d_{r\Delta}(t_1,t_2)=\frac{1}{Z_\Delta}\sum_{\delta_1\in\Delta(t_1)} \min_{\delta_2\in\Delta(t_2)} d_{r\delta}(\delta_1,\delta_2)$}

  where $Z_\Delta=\max\{|\Delta(t_1)|,|\Delta(t_2)|\}$, and $d_{r\delta}$:
%%   where $Z_\Delta=\max\{|\Delta(t_1)|,|\Delta(t_2)|\}$ is a normalization factor, and $d_{r\delta}$ is computed as:

  \myeq{$d_{r\delta}(\delta_1,\delta_2)=\frac{1}{\max\{|\delta_1|,|\delta_2)|\}} \sum_{l_i\in \delta_1} \min_{l_j\in \delta_2} d_{f'}(l_i,l_j)$}

    $d_{f'}(l_i,l_j)=d_{f}(l_i,l_j)$ when the literals correspond to predicates. We use $d_f$ since actions and literals
    in the state ($l_j, l_j$) share the same format (a name and some arguments). However, when they correspond to
    functions, since functions have numerical values, we have to use a different function $d_n$. In this case, each
    $l_i$ will have the form $f_i(\mbox{arg}_i)=v_i$. $f(\mbox{arg}_i)$ has the same format as a predicate (or action)
    with a name $f_i$ and a set of arguments $\mbox{arg}_i$, so we can use $d_f$ on that part. The second part is the
    functions' value. In that case, we compute the absolute value of the difference between the numerical values of both
    functions and divide by the maximum possible difference ($M$) to normalize:\footnote{We use a large constant in practice.}

    \myeq{$d_n(l_i,l_j)= d_f(f_i(\mbox{arg}_i),f_j(\mbox{arg}_j))\times \frac{\mbox{abs}(v_i-v_j)}{M}$}
%%     \[d_n(f_i(C_i)=v_i,f_j(C_j)=v_j)=\\

    We multiply both, since we see the distance on the arguments as a weight that modifies the difference in numerical
    values. For example, if

    \centering $\delta_1$={\tt \{acc-owner(i1,i2),balance(i2)=20\}},\\
    $\delta_2$={\tt \{acc-owner(i1,i3),balance(i3)=10\}},

    \myeq{$d_{r\Delta}(\delta_1,\delta_2)=\frac{1}{2}(\min\{0.25,1\}+\min\{1,0.5\times \frac{|20-10|}{M}\})$}
%%     In case the names of the functions are not the same, $d_{f'}=0$.
  \end{itemize}

  Once we have a distance metric between traces, we use an instance-based technique, as $k$NN, to classify a new trace
  according to the $k$ traces with minimum distance, and computing the mode of those traces' classes. Since the
  classifier takes a trace as input, \cabbot also allows for on-line classification with the current trace up to a given
  simulation step. A nice property of $k$NN is that we can explain how a behavior was classified by pointing out the
  closest previous cases.

\section{Generation of Synthetic Behavior}

\begin{figure}[hbt]
  \placefigure{0.4}{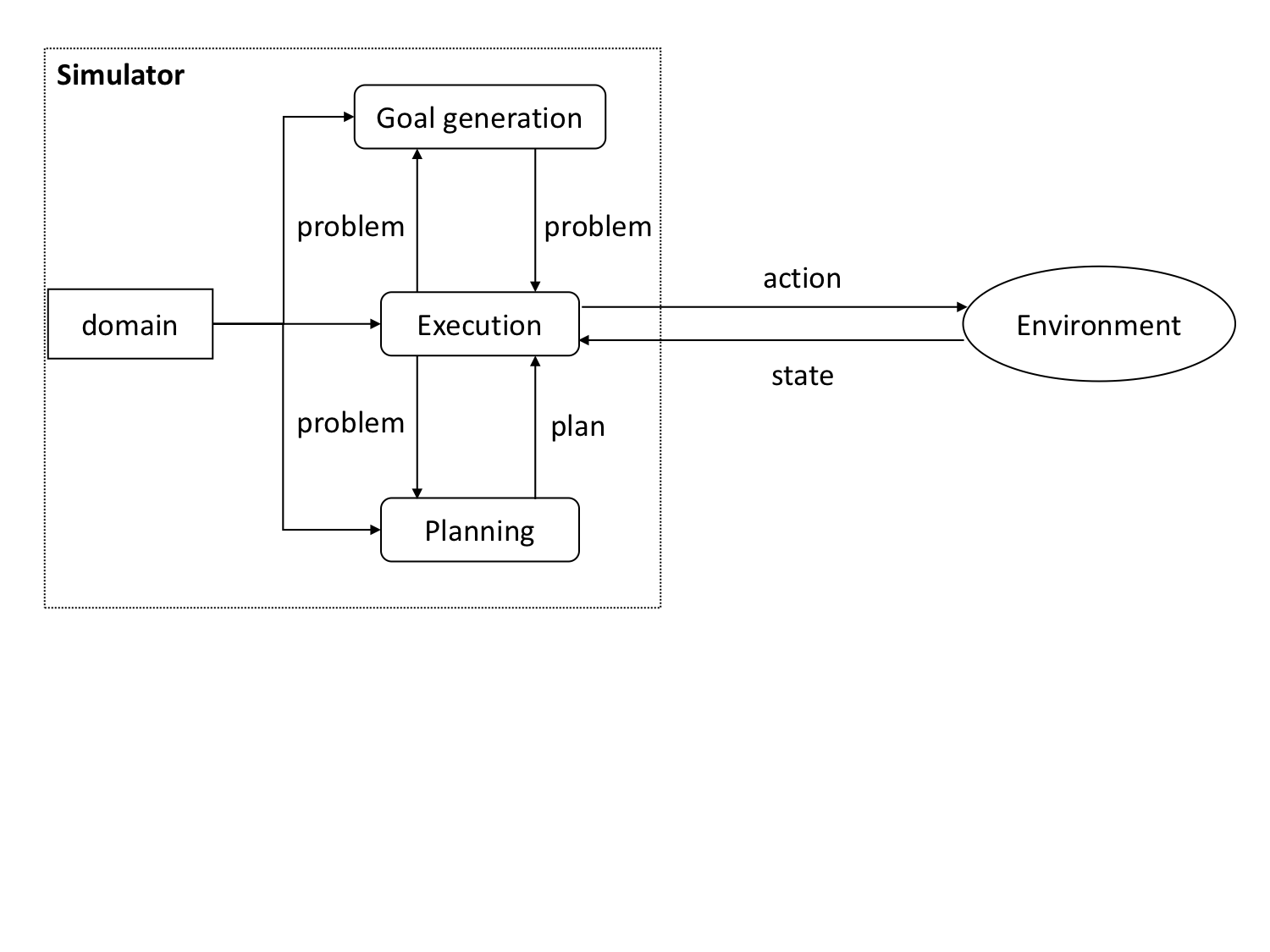}
  \vspace*{-2cm}
\caption{High level view of the simulator.}
\label{fig:arch}
\end{figure}

In real world applications, traces will come from observations of other agents' actions. In this paper, we have also
developed a simulator that can produce those traces for the $\planning$ agent.  Figure~\ref{fig:arch} shows a high level
outline of the simulator. $\planning$ takes actions in the environment by using a rich reasoning model that includes
planning, execution, monitoring and goal generation. It is inspired in some planning and execution
architectures~\cite{spark12}, where the main difference lies on the dynamic generation of goals. In particular, the goal
generation component allows the agent to change or generate new goals on-line as in past work on goal
reasoning~\cite{aicomm18-editorial}.

The components of the simulator for the planning agents are: the {\it Execution}, that takes a domain and problem
description and follows a reasoning cycle that involves generating a new plan by calling {\it Planning}, executing the
next action(s) from the current plan in the environment and observe the next state, and obtaining new goals or state
components from {\it Goal generation}. The simulator is domain independent, except for the Goal reasoning that needs to
generate behavior corresponding to at least two types of agents in the same domain. Now, we present a description of
each module.

\subsection{Execution}

Execution performs several tasks for some iterations:

\begin{itemize}
\item if there is no plan, or there is a reason for replanning, it calls Planning to generate a new plan. Reasons for
  replanning include: the state received from the environment is not the expected one (it does not fully match the state
  predicted by the effects of the most recently action); and Goal generation has returned new goals and/or changes in
  the state. We are using a standard planner for replanning, but it can be substituted by replanning
  algorithms~\cite{lpgfoxetal06,workshop-icaps12-errtplan}.
\item if there is a plan in execution, it selects the next action to execute and sends it to the environment. The
  environment simulates the execution of the action and returns a new state. As mentioned above, the new state can be
  the one defined by the effects (deterministic execution). Our simulator also includes the possibility of defining
  non-deterministic execution of actions, as well as the appearance of exogenous events.
\item at each step, it also calls Goal generation for changes in the goals or partial descriptions of states, as
  explained below.
\item the interaction with the environment also generates a trace of observations that will be used for both training
  and testing of the learning component of $\observer$. As explained before, the trace contains a sequence of actions
  and states from the point of view of $\observer$. Therefore, Execution applies a filter on both so that it only
  includes in the trace its observable elements. Observability is defined for each domain. We opted for a simplified way
  to define it as the sets of lifted actions and predicates that can be observed by $\observer$. Any ground action or
  state literal of a lifted action or predicate on those sets will be observable. Besides, $\observer$ might not see the
  actual executed action but another one (noisy observations). Also, it might not be able to see some of the actions
  even if they are in the observable set (a further aspect of partial observability).
\item each simulation finishes after a predefined number of simulation steps (horizon) that is a parameter, or after a
  plan has not been found in a given time bound. We set the time bound with a low value (10 seconds), since this is
  enough in the experimental domains we have used in most cases.
\end{itemize}

\subsection{Goal generation}

This component allows agents to generate believable behavior whose goals evolve over time depending on the current state
of the environment. It takes as input the current problem description (state, goals and instances) and returns a new
problem description. The first obvious effect of this module is to change goals. In order to do so, we have defined two
kinds of behavior for each domain by changing the goals of each type of behavior. For instance, in the case of a
terrorist domain, we define two types of agents: regular person and terrorist. The regular person would generate goals
of going from one place to another. When the simulator has achieved the previous goal (moving to a place), this module
will generate a new goal of being somewhere else randomly chosen. However, randomly, the knapsack that it carries might
fall down and be forgotten by the person. So, when the person notices that it does not carry the knapsack, it will
generate a new goal to hold it again. In the case of the terrorist, this module will randomly generate the goal of not
carrying the knapsack. And even if it knows that it is not carrying the knapsack, it will not generate as goal to carry
it again, as in the case of the regular person. As a reminder, the observer does not know the goals of the other agent.

This module can also change the problem state and instances. This is useful for generating new components of the state
on-line, as with partial observability of a rich environment. Suppose, we want to simulate an open environment where
agents wander around and go to places that were not defined originally in the initial problem description. One
alternative consists of defining a huge state (and associated instances) in the initial problem description to account
for the whole map. This forces the planner to generate many more instantiations than the ones actually needed to plan in
the first simulation steps. The ability of Goal generation to change the state and instances descriptions, allows the
simulator to generate new parts of the world (or even remove visited ones if not further needed) on the fly, making the
process more efficient and dynamic.

%% \subsection{Environment}
%% 
%% Stochastic environment due to stochastic action execution, as well as exogenous events.

\subsection{Planning}

We are working in a domain-independent setting. Therefore, domain and problem models are specified in PDDL. Thus, any
PDDL complaint planner could be used for this purpose. In particular, we are using some domains with extensive use of
numeric variables (using PDDL functions). So, we are constrained to planners that can reason with numeric preconditions
and effects. Examples of planners we are using are: {\sc lpg}~\cite{LPG}; {\sc cbp}~\cite{ln-caepia10}, or {\sc
  sayphi}~\cite{applied-intelligence-cbr}. As expected, planners take as input a domain and problem description in PDDL,
and return a plan that solves the corresponding planning task. All these planners generate sub-optimal solutions.

\section{Experiments}

We will first describe the experimental setting and then show and analyze the results.

\subsection{Experimental setting}

Due to the lack of existing domains in the planning community that address the task of behavior classification from
planning-execution traces, we have defined several new domains:

\begin{itemize}
\item {\bf terrorist}: a domain where people move around a grid that represents an open place (station, airport, square,
  \ldots) holding a knapsack. Regular people might accidentally drop the knapsack (with probability 0.2), but they try
  to recover it when they find out. Terrorists drop the knapsack (with probability 0.4) and leave it there. The model is
  composed of three actions (move, drop and take) and four predicates. The goal is to classify in terrorist or regular
  behavior from the observed traces. There is full observability in this domain, given that all actions and states are
  observable, and cannot differentiate between the two types of agents.
\item {\bf service cars}: some vehicles move around the streets of a city network. The model comprises seven actions and
  seven predicates. Actions include: moving from one street section to another connected one, boarding and unboarding a
  vehicle, stopping a vehicle and moving it again. The goal is to classify the vehicles that are particular cars from
  the service cars (taxis or equivalent). All actions and predicates are observed, except for two predicates (whether a
  driver of a car owns the car, and whether there is a passenger inside a service car or not). There are two board
  actions depending on the type of vehicle, but the observer cannot differentiate between the two. The same applies to
  debark actions. The probability a new goal related to moving someone appears is 0.6 in the case of service cars, while
  the probability a new goal related to moving the owner appears is 0.2 in case of private cars.
\item {\bf customer journeys (journey)}: customers access the mobile application of a bank and perform several
  operations. The model comprises 22 actions, 24 predicates and 2 functions. Actions include: logging in,
  checking or changing diverse information on their accounts, or performing financial operations. The goal is to
  classify between customers that are active with the application from the ones that do not use it. The observable
  actions and predicates are equal for both types of customers. The main difference is the probability of a goal
  appearing at some point (need of a customer of performing some operation). Active customers will have a higher
  probability than non-active ones.
\item {\bf customer journeys (digital-journey)}: another version of the previous domain, where the task consists of
  classification between digital users and traditional users. In terms of behavior, digital users have a higher
  probability of performing digital-based operations (such as quick payments) and traditional users tend to have a lower
  probability on those operations, but a higher one on traditional operations (such as paying bills).
\item {\bf anti-money laundering (AML)}: customers of a financial institution perform operations such as money
  transfers, payments, or deposits. In the meantime, not observable by $\observer$, these customers are either involved
  in criminal activities, or are regular customers. The challenge in this domain consists of characterizing the type of
  behavior from observations related to standard activities with the bank. The model comprises 33 actions, 37 predicates
  and 12 functions. Actions include: criminal activities, getting a job, getting a payroll, or making financial
  operations. The goal consists of classifying between money laundering individuals and regular
  individuals. Observability is restricted to the information that a bank can have on a given customer. Therefore,
  predicates as someone being a criminal or getting dirty money are not observable, while predicates related to making
  transactions and opening accounts are. We have tried to make this domain rich in terms of the different traces
  generated by the simulator. Therefore, we have defined several probability distributions that affect issues such as
  probabilities of selecting different money laundering strategies by criminals, or buying different kinds of items by
  criminals and regular customers.
\end{itemize}

\no{Other datasets are not useful in this context, where we assume a domain model. For instance, in
  \url{http://odds.cs.stonybrook.edu} there are many datasets related to outlier detection tasks where data consists of
  a set of traces, where each element of the trace includes several features. Examples range from network attacks to
  sentiment analysis of tweets or protein sequence classification. We could not find a good mapping between these sets
  and our task.}

For each domain, we have randomly generated 10 traces of each type of behavior for training and 20 for test (where
classes are uniformly randomly selected). We measure the accuracy of the prediction. We have used $k=1$ for the
experiments, given that we already obtained good results with that value. We have varied the following parameters to see
the impact they have on the results:

\begin{itemize}
\item length of the traces (simulation horizon). We used the values: 5, 10, 20, 50 and 100. Default is 50.
\item similarity function. We have used the defined ones: $d_a$, $d_\Delta$, $d_g$ and $d_r$. Default is $d_r$.
\item probability-goal-appears. We have defined a probability that a set of goals appear at a given time step. Once a
  set of goals appears, it might take several time steps to execute the plan to achieve all goals. In the meantime, we
  do not generate new goals, though the simulator is ready to work with that case too. We used the values 1.0, 0.8, 0.5,
  0.1, 0.05 and 0.01. Default is 1.0.
%% \item probability-random. In order to also compare against random behavior, we used an stochastic planner (the
%%   stochastic version of {\sc cbp}) that takes a value for the desired randomness. At each step, with a
%%   probability-random, it uses a random successor action, and otherwise it uses the best action according to the
%%   heuristic. We have used the values: 0.0 (no randomness), 0.2, 0.5, and 0.8. Default is 0.2.
%% \item probability-noise. At each executed action, $O$ does not observe the real executed action (in case it is
%%   observable) with that probability. Instead, it observes a random action. With the remaining probability, it observes
%%   the action correctly. This only affects the performance of \cabbot if we are using action based distance, since the
%%   observation on the states is not affected. \todo{I could also try to change randomly the state.} We have used the
%%   values: 0.0 (correct observability), 0.4, and 0.8. Default is 0.0.
%% \item probability-observation. At each executed action, $O$ observes the action with that probability. Otherwise, it
%%   observes nothing. We have used the values: 0.2, 0.6 and 1.0 (full observability). Default is 1.0.
%% \item off-line vs. on-line classification. We can observe the whole trace and classify it, or we can perform on-line
%%   classification afer each observation together with the previous ones.
\end{itemize}

We will present results by varying these parameters one at a time to observe the impact they have on the performance of
\cabbot.

\subsection{Results}

Table~\ref{tab:prob-goal} shows the results for the {\tt journey} domain. In this domain, the behavior depends on the
probability of a goal arriving for both types of customers: active and non-active. We varied those probabilities to
analyze how their values affect the accuracy of \cabbot. We can observe that when the difference between the two
probabilities gets smaller, the behavior becomes more similar (in terms of activity level of customers) and accuracy of
classification degrades. In the extreme, when the two probabilities are equal -- (0.5, 0.5) case --, the classification
accuracy is equivalent to a random classification (0.55). We will use the combinations $\langle 0.8, 0.01\rangle$ (named
journey-B for bigger difference) and $\langle 0.5, 0.1\rangle$ (named journey-S for smaller difference) for the
remaining comparisons.

\begin{table}[hbt]
  \begin{center}
  \begin{tabularx}{0.6\textwidth}{c *{4}{Y}}
    \toprule
    & \multicolumn{4}{c}{\bf Prob. non-active}\\
    \multicolumn{1}{c}{\bf Prob. active} & \multicolumn{1}{c}{\bf 0.01} &
    \multicolumn{1}{c}{\bf 0.05} & \multicolumn{1}{c}{\bf 0.1} & \multicolumn{1}{c}{\bf 0.5}\\
    \midrule
    {\bf 0.5} & 1.00 & 1.00 & 0.85 & 0.55\\
    {\bf 0.8} & 1.00 & 0.95 & 0.85 & 0.70\\
    {\bf 1.0} & 1.00 & 1.00 & 0.95 & 0.80\\
%% they were using knn instead of ribl which does not normalize the traces when learning    
%%     {\bf 0.5} & 0.70 & 0.65 & 0.55\\
%%     {\bf 0.8} & 0.95 & 0.55 & 0.35 \\
%%     {\bf 1.0} & 0.75 & 0.70 & 0.35 \\
    \bottomrule
\end{tabularx}
\end{center}
  \caption{Classification accuracy in the customer journey domain varying the probability of appearing goals for the two
    kinds of customers, active and non-active.}
  \label{tab:prob-goal}
\end{table}

The next results of the experiments are presented in Table~\ref{tab:horizon}. Rows represent the domains, and the
columns are different lengths of the traces (horizons). The values correspond to the accuracy of \cabbot fixing all
other parameters to their default values. The results show that \cabbot is able to correctly classify behavior traces in
a high percentage of cases.  We observe that we do not need a high number of traces nor lengthy traces to obtain good
results. As expected, \cabbot had less accuracy in shorter traces, since it has observed less number of actions/states,
so it is harder to correctly classify the behavior. In the case of the {\tt journey} domain, the longer traces allow for
more goals to appear in the case of non-active customers, making the classification harder. Also, as it was observed
before, the results with a smaller difference of probability values are worse than with a bigger difference, specially
in the case of shorter traces' lengths.

\begin{table}[hbt]
  \begin{center}
  \begin{tabularx}{0.6\textwidth}{c *{5}{Y}}
    \toprule
    & \multicolumn{5}{c}{\bf Length of traces}\\
    \multicolumn{1}{c}{\bf Domain} & \multicolumn{1}{c}{\bf 5} &
    \multicolumn{1}{c}{\bf 10} & \multicolumn{1}{c}{\bf 20} &
    \multicolumn{1}{c}{\bf 50} & \multicolumn{1}{c}{\bf 100}\\
    \midrule
    terrorist & 0.60 & 0.90 & 0.95 & 1.00 & 1.00\\
    service car & 0.60 & 0.95 & 1.00 & 1.00 & 1.00\\
    journey-B & 1.00 & 0.95 & 0.85 & 0.80 & 0.95\\
    journey-S & 0.45 & 0.80 & 0.60 & 0.95 & 0.85\\
    digital-journey & 0.75 & 0.85 & 0.90 & 1.00 & 1.00 \\
    AML & 1.00 & 1.00 & 1.00 & 0.90 & 0.95\\
    %% with knn
%%     terrorist & 0.60 & 0.90 & 0.95 & 1.00 & 1.00\\
%%     service car & 0.60 & 0.95 & 1.00 & 1.00 & 1.00\\
%%     journey-B & 1.0 & 0.95 & 0.85 & 0.80 & 0.90\\
%%     journey-S & 0.5 & 0.75 & 0.70 & 0.60 & 0.55\\
%%     AML & 1.00 & 1.00 & 1.00 & 0.90 & 0.95\\
    \bottomrule
\end{tabularx}
\end{center}
  \caption{Classification accuracy in different domains varying the length of the trace.}
  \label{tab:horizon}
\end{table}

Table~\ref{tab:similarity} shows the results when we vary the similarity function. As we can see, the accuracy is
perfect in most cases for all domains except for the customer journeys one. Even if the intention when generating the
two kinds of behavior was to include slight differences, the learning system is able to detect those by using the
different similarity functions. In the case of the {\tt journey} domain, we can see that the actions-based distance
obtains better results than the one based on comparing goals. Since this domain has many different goals, when goals
appear the traces differ more on the goals than on the actions achieving the goals. Also, the similarity function used
does not affect much in this domain to differentiate between bigger (B) or smaller (S) probability differences.

\begin{table}[hbt]
\begin{center}
  \begin{tabularx}{0.6\textwidth}{c *{4}{Y}}
    \toprule
    & \multicolumn{4}{c}{\bf Similarity function}\\
    \multicolumn{1}{c}{\bf Domain} & \multicolumn{1}{c}{\bf $d_a$} &
    \multicolumn{1}{c}{\bf $d_\Delta$} & \multicolumn{1}{c}{\bf $d_g$}
     & \multicolumn{1}{c}{\bf $d_r$}\\
    \midrule
    terrorist & 1.00 & 1.00 & 0.95 & 0.90\\
    service car & 0.50 & 1.00 & 1.00 & 1.00\\
    journey-B & 1.00 & 0.50 & 1.00 & 0.80 \\
    journey-S & 0.75 & 0.50 & 1.00 & 0.95 \\
    digital-journey & 1.00 & 0.95 & 1.00 & 1.00\\
    AML & 1.00 & 1.00 & 1.00 & 1.00\\
    \bottomrule
\end{tabularx}
\end{center}
  \caption{Classification accuracy in different domains varying the similarity function.}
  \label{tab:similarity}
\end{table}

\cabbot can make on-line classification of traces as soon as observations are made. Table~\ref{tab:online} shows the
average number of observations before making the final classification decision when varying the similarity
function. While in the {\tt AML} and {\tt service car} domains, it takes a small number of steps to make the final
decision, the number of steps required in the other two domains is higher. This is specially true in the case of the
{\tt journey} domain for the same reasons discussed above; i.e. goals could take some time to appear.

\begin{table}[hbt]
\begin{center}
  \begin{tabularx}{0.6\textwidth}{c *{4}{Z}}
    \toprule
    & \multicolumn{4}{c}{\bf Similarity function}\\
    \multicolumn{1}{c}{\bf Domain} & \multicolumn{1}{c}{\bf $d_a$} &
    \multicolumn{1}{c}{\bf $d_\Delta$} & \multicolumn{1}{c}{\bf $d_g$}
     & \multicolumn{1}{c}{\bf $d_r$}\\
    \midrule
    terrorist & 4.20 & 6.40 & 26.40 & 15.90\\
    service car & 0.00 & 2.90 & 26.20 & 0.70\\
    journey-B & 15.90 & 7.30 & 17.90 & 2.30\\
    journey-S & 25.00 & 25.00 & 16.40 & 11.10\\
    digital-journey & 2.80 & 10.60 & 10.55 & 5.90\\
    AML & 2.30 & 1.40 & 5.25 & 1.60\\
    \bottomrule
\end{tabularx}
\end{center}
\caption{Average number of observations before making the final classification decision when varying the similarity
  function in several domains.}
  \label{tab:online}
\end{table}

Table~\ref{tab:observability} shows the results when we vary the probability of partial observability. We can see that
when the probability of making an observation at a given time step decreases, so does the accuracy of the learning
system and correspondingly the number of steps it takes the learning system to converge to the final classification
increases. In the extreme, when the probability is 0.01 for a length of history of 50, the traces will at most consist
of one or two elements, so classifying the traces becomes a hard task as shown by the low probabilities. The rate at
which the accuracy decreases varies across domains. In the case of {\tt AML}, {\tt digital-journey} and {\tt journey-B}
domains, there is a slow decrease in accuracy. In the other three domains, the drop in accuracy is more acute starting
at even a probability of observation of 0.5 in the {\tt terrorist} domain.

\begin{table}[hbt]
\begin{center}
  \begin{tabularx}{0.8\textwidth}{c *{4}{Y}}
    \toprule
    & \multicolumn{4}{c}{\bf Probability of partial observations}\\
    \multicolumn{1}{c}{\bf Domain} & \multicolumn{1}{c}{\bf 1.0} &  \multicolumn{1}{c}{\bf 0.5} &
    \multicolumn{1}{c}{\bf 0.1} & \multicolumn{1}{c}{\bf 0.01}\\
    \midrule
    terrorist & 0.95 & 0.45 & 0.50 & 0.50\\
    service car & 1.00 & 1.00 & 0.85 & 0.45\\
    journey-B & 0.90 & 0.90 & 1.00 & 0.60\\
    journey-S & 0.85 & 0.85 & 0.80 & 0.45\\
    digital-journey & 0.90 & 0.75 & 0.55 & 0.45\\
    AML & 1.00 & 0.90 & 0.80 & 0.35\\
%%     terrorist & 0.90 (13.8) & 0.55 (21.8) & 0.35 (14.4) & 0.40 (0.5)\\
%%     service car & 1.00 (0.6) & 0.90 (12.3) & 0.65 (15.4) & 0.55 (4.1)\\
%%     journey-B & 1.00 (0.5) & 0.85 (5.6) & 0.95 (10.0) & 0.40 (0.0)\\
%%     journey-S & 0.85 (16.8) & 0.85 (6.3) & 0.50 (1.6) & 0.50 (0.0)\\
%%     digital-journey & 0.95 (6.7) & 0.90 (5.9) & 0.85 (12.6) & 0.45 (4.4)\\
%%     AML & 1.00 (3.0) & 0.90 (4.1) & 0.85 (11.0) & 0.50 (0.0)\\
    \bottomrule
\end{tabularx}
\end{center}
\caption{Classification accuracy in different domains varying the probability of partial observability.}
%%  In parenthesis, the number of steps until it converges to the final decision.
  \label{tab:observability}
\end{table}

Table~\ref{tab:failure} shows the results when we vary the probability of an execution failure of individual action
(degree of non-determinism). When an action fails, it stays in the same state. Since the length of the history is 50
steps, even if some actions fail, \cabbot is still getting enough observations to make accurate classifications.

\begin{table}[hbt]
\begin{center}
  \begin{tabularx}{0.8\textwidth}{c *{3}{Y}}
    \toprule
    & \multicolumn{3}{c}{\bf Probability of execution failure}\\
    \multicolumn{1}{c}{\bf Domain} & \multicolumn{1}{c}{\bf 0.0} &  \multicolumn{1}{c}{\bf 0.2} &
    \multicolumn{1}{c}{\bf 0.4}\\
    \midrule
    terrorist & 0.95 & 0.90 & 0.80\\
    service car & 1.00 & 1.00 & 1.00\\
    journey-B & 0.90 & 0.95 & 0.95 \\
    journey-S & 0.90 & 0.95 & 0.80\\
    digital-journey & 0.70 & 0.85 & 0.75\\
    AML & 1.00 & 1.00 & 1.00\\
%%     terrorist & 0.90 (13.8) & 0.90 (8.5) & 0.85 (9.8)\\
%%     service car & 1.00 (0.6) & 1.00 (1.0) & 1.00 (0.8)\\
%%     journey-B & 1.00 (0.5) & 0.90 (2.7) & 0.75 (7.0) \\
%%     journey-S & 0.85 (16.8) & 0.80 (17.9) & 0.80 (15.1)\\
%%     digital-journey & 0.95 (6.7) & 0.60 (2.5) & 0.75 (14.6)\\
%%     AML & 1.00 (3.0) & 1.00 (2.2) & 1.00 (2.6)\\
    \bottomrule
\end{tabularx}
\end{center}
\caption{Classification accuracy in different domains varying the probability of individual action execution failure. In
  parenthesis, the number of steps until it converges to the final decision.}
  \label{tab:failure}
\end{table}

%% \todo{I do not know if this makes any sense at all} Table~\ref{tab:noise} shows the results when we vary the probability
%% of noisy observations. We can see that the increase of noise does not necessarily decrease the accuracy. Since \ribl is
%% based on computing similarities between traces, the higher the noise, the higher the diversity among traces is. \todo{it
%%   is not a good explanation} In three domains (terrorist, journey-B and digital-journey) the accuracy decreases with
%% higher levels of noise (from 0.0 to 0.2), but it increases again with a noise level of 0.4.
%% 
%% \begin{table}[hbt]
%% \begin{center}
%%   \begin{tabularx}{0.45\textwidth}{c *{3}{Y}}
%%     \toprule
%%     & \multicolumn{3}{c}{\bf Probability of noisy observations}\\
%%     \multicolumn{1}{c}{\bf Domain} & \multicolumn{1}{c}{\bf 0.0} &  \multicolumn{1}{c}{\bf 0.2} &
%%     \multicolumn{1}{c}{\bf 0.4}\\
%%     \midrule
%%     terrorist & 0.95 & 0.65 & 0.80\\
%%     service car & 1.0 & 1.0 & 1.00\\
%%     journey-B & 1.0 & 0.55 & 0.95\\
%%     journey-S & 0.8 & 0.85 & 0.90\\
%%     digital-journey & 0.85 & 0.6 & 0.85\\
%%     AML & 1.0 & 1.0 & 1.00\\
%%     \bottomrule
%% \end{tabularx}
%% \end{center}
%% \caption{Classification accuracy in different domains varying the probability of noisy observations.}
%%   \label{tab:noise}
%% \end{table}

\section{Related work}

Given some sequence of events, there have been several learning tasks defined: sequence prediction (what the next step
is going to be)~\cite{bernard2019accurate}; sequence generation (learning to generate new sequences, e.g. simulation);
sequence recognition (determine whether the sequence is legitime or belongs to a given type); sequential decision making
(how to make decisions over time, e.g. planning). This paper deals with sequence recognition or classification.

This task has been addressed by using different types of techniques~\cite{xing2010brief} based on: features, distances
or models. Features can be the presence or frequency of $k$-grams for all grams of size $k$.
%% In our case, generating $k$-grams grows very fast exponentially (if we consider all actions groundings) or not
%% useful (if we consider only the names of actions). \todo{though our approach is based on names of actions and works
%%   well}
Model-based assumes an underlying probabilistic model and learns the parameters (Naive Bayes, HMM, ...). In our case,
the number of symbols in the alphabet is huge (if groundings), so computing conditional probabilities is intractable, or
is very small (action schemas) and probably not useful. Otherwise, we would have to rely on domain knowledge to know,
for instance, that the transaction amounts (not part of the actions) are relevant, or the sum of amounts of several
consecutive transactions. So, we have opted to use a distances-based approach.  Our learning task is also related to
detecting anomalous behavior or outliers detection~\cite{chandola2010anomaly,gupta2013outlier} where the techniques are
the same ones.  The main difference with respect to previous work is that their definition of traces is very simplistic
in most cases: small number of action labels; no representation of state nor goals; and they do not handle relational
data.  From the point of view of classical automated planning, there has been related work on goal/plan
recognition~\cite{ramirez10}. However, as we discussed in the introduction, the task we deal with here is not about
predicting the goal/plan, but about classifying a given behavior in a set of behavior classes.

We have used several similarity functions such as the ones based on Jaccard distance or \ribl. Other similarity
functions have been defined in related tasks, such as process mining~\cite{becker2012comparative}, \change{plan
  diversity~\cite{RobertsHR14} or plan stability~\cite{lpgfoxetal06} (see~\cite{ontanon2020overview} for an extensive
  review). These previous similarity functions used mainly the actions in the plan, but did not include the
  corresponding states.}

Some of our domains have been analyzed previously by similar approaches: understanding customer journeys in the field of
marketing~\cite{Lemon2016}; predicting an on-line buy action from the sequence of clicks~\cite{Bertsimas03}; process
mining~\cite{ProcessMining2016}; intrusion detection in a computer network or system~\cite{Scholau}; or anti-money
laundering~\cite{lopezrojas}. None of them used a representation based on planning tasks, nor any relational learning
approach. So, their approaches relied on carefully selecting the features to be used for defining the learning examples.

\section{Conclusions}

We have presented four main contributions. The first contribution consists of posing the sequence classification task in
terms of a richer representation framework than previous work. We use goals, states and actions to include the traces
rationale in the traces description. The second contribution consists of a learning technique that takes a set of
training traces of other agents's behavior and can classify later traces in different classes. The third contribution is
a simulator that generates synthetic behaviors where agents can dynamically change their goals, and therefore their
plans. Execution of those plans is stochastic, so those agents are able to monitor the execution and replan when
needed. Finally, the fourth contribution is a set of automated planning domains that can be used for comparison in
future work. Experimental results show that this approach performs well in some domains, including variations of real
finance-related domains.

\section*{Acknowledgements}

\change{This paper was prepared for information purposes by the Artificial Intelligence Research group of JPMorgan Chase
  \& Co. and its affiliates (``JP Morgan''), and is not a product of the Research Department of JP Morgan.  JP Morgan
  makes no representation and warranty whatsoever and disclaims all liability, for the completeness, accuracy or
  reliability of the information contained herein.  This document is not intended as investment research or investment
  advice, or a recommendation, offer or solicitation for the purchase or sale of any security, financial instrument,
  financial product or service, or to be used in any way for evaluating the merits of participating in any transaction,
  and shall not constitute a solicitation under any jurisdiction or to any person, if such solicitation under such
  jurisdiction or to such person would be unlawful. The authors thank Alice Mccourt for her useful revision of the
  paper. The authors would like to thank Sameena Shah for her discussions on the applications of this work to the AML
  task. \textcopyright 2020 JPMorgan Chase \& Co. All rights reserved}

\bibliographystyle{named}
%% \bibliography{/Users/dborrajo/papers/bib/general-daniel,/Users/dborrajo/papers/bib/daniel}

\end{document}